\documentclass[letterpaper]{article} 
\usepackage{aaai2026}  
\usepackage{times}  
\usepackage{helvet}  
\usepackage{courier}  
\usepackage[hyphens]{url}  
\usepackage{graphicx} 
\usepackage{natbib}  
\usepackage{caption} 
\usepackage{algorithm}
\usepackage{algorithmic}

\usepackage{newfloat}
\usepackage{listings}
\DeclareCaptionStyle{ruled}{labelfont=normalfont,labelsep=colon,strut=off} 
\floatstyle{ruled}
\newfloat{listing}{tb}{lst}{}
\floatname{listing}{Listing}
\title{Seeing is Believing: Rich-Context Hallucination Detection \\ for MLLMs via Backward Visual Grounding}

\makeatletter
\newcommand{\sharedfootnote}[2]{%
  \begingroup
  \renewcommand\thefootnote{#1}%
  \footnotetext{#2}%
  \endgroup
}
\makeatother

\author{
    Pinxue Guo\textsuperscript{\rm 1}$^*$,
    Chongruo Wu\textsuperscript{\rm 3}$^*$,
    Xinyu Zhou\textsuperscript{\rm 2},
    Lingyi Hong\textsuperscript{\rm 2},
    Zhaoyu Chen\textsuperscript{\rm 1},\\
    Jinglun Li\textsuperscript{\rm 1},
    Kaixun Jiang\textsuperscript{\rm 1},
    Sen-ching Samson Cheung\textsuperscript{\rm 4},
    Wei Zhang\textsuperscript{\rm 2}$^\dagger$,
    Wenqiang Zhang\textsuperscript{\rm 1,2}$^\dagger$
}
\affiliations{
    \textsuperscript{\rm 1}College of Intelligent Robotics and Advanced Manufacturing, Fudan University\\
    \textsuperscript{\rm 2}College of Computational Science and Artificial Intelligence, Fudan University\\
    \textsuperscript{\rm 3}Independent Researcher ~
    \textsuperscript{\rm 4}Electrical and Computer Engineering, University of Kentucky
}

\usepackage{bibentry}

\usepackage{multirow}
\usepackage{booktabs}
\usepackage{xcolor}
\usepackage{comment}
\usepackage{amsmath}
\usepackage{amssymb}

\usepackage{hhline}
\usepackage{colortbl}
\usepackage[table]{xcolor}

\usepackage{makecell}
\usepackage{multicol}
\usepackage{soul}

\usepackage{tikz}
\usepackage{graphicx}
\usepackage{calc}

\newcommand{\aaainew}[1]{\textcolor{black}{#1}}
\newcommand{\samson}[1]{\textcolor{black}{#1}}

\newcommand{\modelname}{{\fontfamily{cmtt}\selectfont VBackChecker}}
\newcommand{\ourbenchmark}{{\fontfamily{cmtt}\selectfont R$^2$-HalBench}}
\newcommand{\ourinstructdata}{{\fontfamily{cmtt}\selectfont R-Instruct}}

\begin{document}

\maketitle

\begin{tikzpicture}[remember picture,overlay,shift={(current page.north west)}]
\node[anchor=north west, xshift=3.17cm, yshift=-3.2cm]{\scalebox{1}[1]{\includegraphics[width=1.05cm]{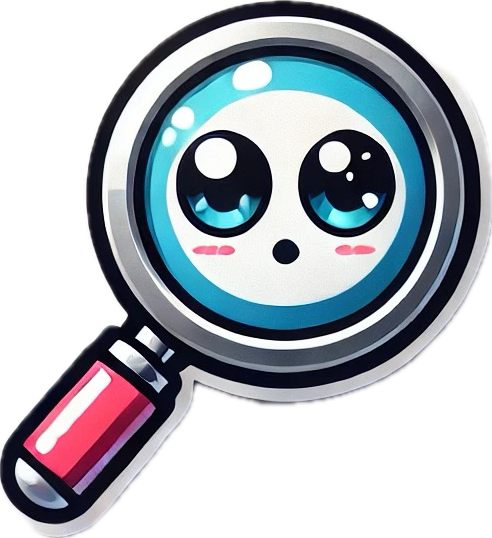}}};
\end{tikzpicture}

\sharedfootnote{*}{Equal contribution.} \sharedfootnote{$\dagger$}{Corresponding authors.}

\begin{abstract}

Multimodal Large Language Models (MLLMs) have unlocked powerful cross-modal capabilities, but still significantly suffer from hallucinations. 
As such, accurate detection of hallucinations in MLLMs is imperative for ensuring their reliability in practical applications.
To this end, guided by the principle of “Seeing is Believing”, we introduce \modelname{}, a novel \aaainew{reference-free} hallucination detection framework that verifies the consistency of MLLM-generated responses with visual inputs, by leveraging a pixel-level Grounding LLM equipped with reasoning and referring segmentation capabilities.
This reference-free framework not only effectively handles rich-context scenarios, but also offers interpretability.
\aaainew{To facilitate this}, an innovative pipeline is accordingly designed for generating instruction-tuning data \aaainew{(\ourinstructdata)}, featuring rich-context descriptions, grounding masks, and hard negative samples.
We further establish \ourbenchmark{}, a new hallucination benchmark for MLLMs, which, unlike previous benchmarks, encompasses real-world, rich-context descriptions from 18 MLLMs with high-quality annotations, spanning diverse object-, attribute-, and relationship-level details. 
\modelname{} outperforms prior complex frameworks and achieves state-of-the-art performance on \ourbenchmark{}, even rivaling GPT-4o's capabilities in hallucination detection. It also surpasses prior methods in the pixel-level grounding task, achieving over a 10\% improvement. 
All codes, data, and models are available at https://github.com/PinxueGuo/VBackChecker.
\end{abstract}
    
\vspace{-3mm}
\section{Introduction}

\begin{figure}[t]
    \centering
    \includegraphics[width=0.9\linewidth]{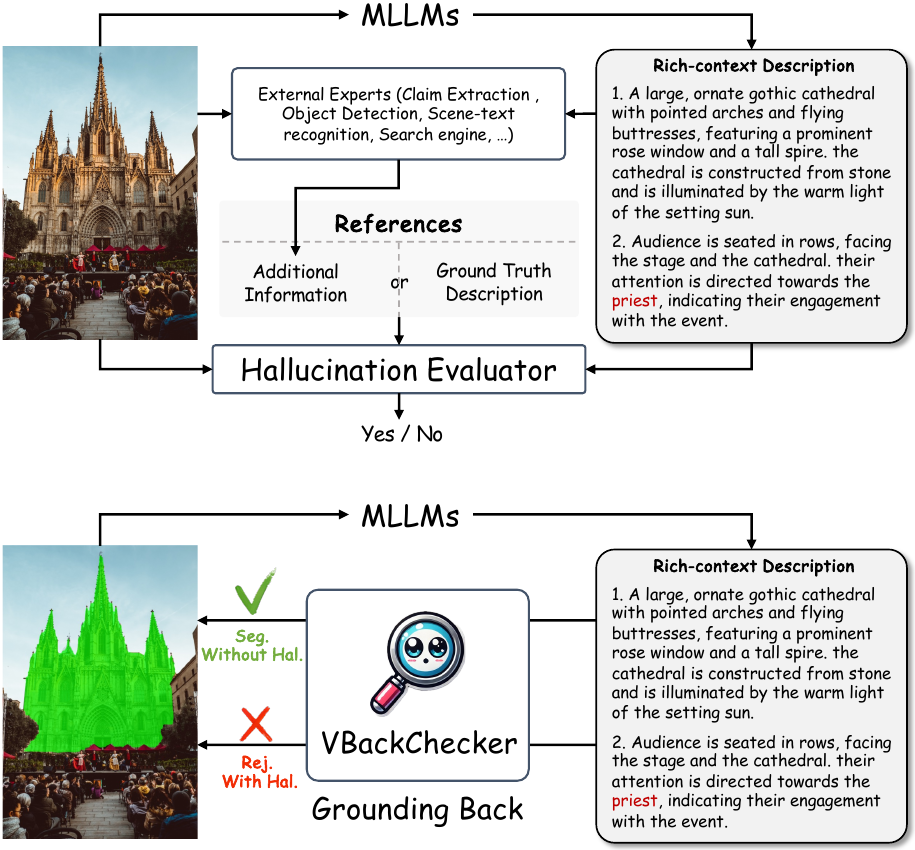}
    \caption{
    Illustration of our proposed \modelname{}. Unlike previous methods (Top), our \modelname{} (Bottom) identifies hallucination in MLLMs by leveraging a pixel-level grounding approach for backward verification, 
    without any references or external experts.
    }
    \label{fig:teaser}
    \vspace{-3mm}
\end{figure}

Multimodal Large Language Models (MLLMs)~\cite{zhu2023minigpt, liu2023visual, alayrac2022flamingo, li2024llava, wang2024qwen2, chen2024internvl, panagopoulou2024x} have recently exhibited remarkable advancements in cross-modal understanding and reasoning, as demonstrated in tasks such as image captioning, visual question answering, and multimodal reasoning.
However, their capabilities remain 
\samson{substantially}
compromised by visual hallucination~\cite{liu2024survey, bai2024hallucination} where the generated content conflicts with the visual input. 
Since this hallucination issue significantly hinders MLLMs' reliability and practical adoption, accurate detection of hallucinations in MLLM-generated response has become an essential priority.
Advanced MLLMs typically generate rich-context responses to accurately describe image content, reflecting the richness of visual information captured in the saying “a picture is worth a thousand words”.
To this end, this paper aims to detect hallucinations in the rich-context responses generated by MLLMs based on the provided image input.

\begin{table*}[t]

\newcommand{\greentick}{\Large\textcolor{green!70!black}{$\checkmark$}}
\newcommand{\redcross}{{\textcolor{red}{$\times$}}}

\begin{center}

\scalebox{1}{

\begin{tabular}{@{\hspace{-1cm}}c@{\hspace{2.1cm}}c@{\hspace{0cm}}}

\begin{minipage}{0.42\textwidth}

\scalebox{0.59}{
\begin{tabular}{c@{\hspace{0.5mm}}c@{\hspace{0.5mm}}c@{\hspace{0.5mm}}cccc}

\toprule
\multirow{2}{*}{ \raisebox{-12pt}{\textbf{\makecell{Evaluation \\Method}}} } & 
\multirow{2}{*}{ \raisebox{-12pt}{\makecell{\textbf{Reference} \\ \textbf{Free}} } } & 
\multirow{2}{*}{ \raisebox{-12pt}{\textbf{\makecell{Without \\External \\Expert}} } } & 
\multirow{2}{*}{ \raisebox{-12pt}{\textbf{\makecell{Without \\Calling \\GPT API}} } } & 
\multicolumn{2}{c}{ \raisebox{-2pt}{\textbf{Explainability}} } & 
\multirow{2}{*}{ \raisebox{-12pt}{\textbf{\makecell{End2End \\Single Model}}} } \\ [0.3\normalbaselineskip] 

\cmidrule{5-6}

& & & & 
\raisebox{-3pt}{\textbf{Visual}} & 
\raisebox{-3pt}{\textbf{Language}} & \\ [0.5\normalbaselineskip] 

\midrule
HaELM~\cite{haelm}
&  
&  \greentick  
&  
&  &  
& \greentick  \\

AMBER~\cite{amber}
&  
& 
& \greentick  
&  &  
&  \\

UniHD~\cite{chen2024unified}
&   
& 
& 
&  &\greentick  
&  \\

GAVIE~\cite{liu2023mitigating}
&   
&  
& 
&  &\greentick  
&  \\

FaithScore~\cite{jing2023faithscore}
& \greentick  
& \greentick    
& 
&  &  
&  \\

\midrule

\modelname{} (Ours)
& \greentick  
& \greentick   
& \greentick   
&\greentick  &\greentick  
&\greentick  \\

\bottomrule
\end{tabular}
}
\label{tab:table_1_methods}
\end{minipage}

&
\begin{minipage}{0.42\textwidth}

\renewcommand{\arraystretch}{1.45}

\scalebox{0.545}{

\begin{tabular}{c@{\hspace{2.5mm}}c@{\hspace{0.5mm}}ccc@{\hspace{2.5mm}}c@{\hspace{2.5mm}}c}
\toprule

\raisebox{-8pt}{\textbf{Benchmark}} & 
\raisebox{-8pt}{\makecell{\textbf{Hallucination} \\ \textbf{Type}}} & 
\raisebox{-8pt}{\makecell{\textbf{Source} \\ \textbf{MLLMs}}} & 
\raisebox{-8pt}{\makecell{\textbf{Real} \\ \textbf{Dis}}} & 
\raisebox{-8pt}{\makecell{ \textbf{Rich-Context} \\ \textbf{Description}}} & 
\raisebox{-8pt}{\makecell{ \textbf{Description}\\ \textbf{Avg Len}}} &
\raisebox{-8pt}{\makecell{\textbf{Max} \\ \textbf{Len}}} \\ [1.8\normalbaselineskip]

\midrule

POPE~\cite{pope} & Obj 
& 0 &  
&  & 2.2 & 3 \\

CIEM~\cite{hu2023ciem} & Obj, Attr
& 1 &
&  & - & - \\

AMBER~\cite{wang2023amber} & Obj, Attr, Rel
& 0 & 
&  & 2.4 & 9 \\

MHaluBench~\cite{chen2024unified} & Obj, Attr, Text 
& 5 & 
& \greentick  
& 13.0 & 58 \\ 

\midrule

\ourbenchmark{} (Ours) & Obj, Attr, Rel 
& 18 
& \greentick 
& \greentick  
& 16.4 & 110 \\ [-0mm]

\bottomrule
\end{tabular}
}
\label{tab:table_1_dataset}
\end{minipage}

\end{tabular}
}
\end{center}

\vspace{-3mm}
\caption{
\textbf{(Left) Comparison of Hallucination Detection Methods}.
\textbf{(Right) Comparison of Hallucination Detection Benchmarks}. “Obj, Attr and Rel" represent abbreviations for Object, Attribute and Relation, respectively. 
“Source MLLMs" denotes the number of MLLMs employed to generate response, 
with “Real Dis" indicating whether responses 
\samson{reflect}
the real distribution. 
The “Description Avg/Max len" represents the avg/max 
\samson{response lengths}, measured in word count.
}
\label{tab:table_1_final}
\vspace{-4mm}
\end{table*}

\samson{Existing}
hallucination detection methods fall into two main categories: reference-based and reference-free approaches. Typically, reference-based methods directly leverage groundtruth text as the comparative reference, where large language models evaluate the generated text against the groundtruth text to identify hallucinations~\cite{wang2023evaluation}, or utilize external expert models like dense object captioning or object detection models, to provide additional references extracted from the image~\cite{liu2023mitigating, wang2023amber, chen2024unified}.
However, reference-based methods rely on annotated GT or external experts, limiting their ability to evaluate unannotated inputs or address capability bottlenecks.
On the other hand, reference-free methods offer more flexibility. A recent work, FaithScore~\cite{jing2023faithscore}, leverages visual question-answering ability to identify hallucinations without references. However, its binary (yes/no) response 
\samson{is susceptible}
to bias and lacks interpretability, restricting its detection capabilities. 
Additionally, since 
\samson{relying on}
LLMs to decompose the input into atomic units\samson{, which }introduces computational overhead and error propagation, 
\samson{further limiting its}
capacity for handling rich-context descriptions.

In this work, we present the Verification by Vision Back (\modelname{}), a \samson{novel} framework illustrated in Fig.~\ref{fig:teaser} directly addresses the challenges of detecting hallucination in MLLMs' responses, by leveraging a pixel-level grounding approach (e.g. LISA~\cite{lai2023lisa}) with reasoning and referring segmentation capabilities. 
The key motivation behind our framework is straightforward:\textbf{ Seeing is Believing}. 
If the element described in the generated response could be accurately grounded back to the image, there is no hallucination; otherwise, a hallucination is present. 
Against previous methods, shown in Table~\ref{tab:table_1_final}~(Left), \modelname{} is a reference-free 
\samson{framework that can handle}
arbitrary inputs 
\samson{and provide interpretable-outputs}
on both vision and language modalities, without external expert assistance.

Towards facilitating the development of \modelname{}, we introduce two key innovations that substantially enhance hallucination detection capabilities. 
First, 
to equip \modelname{} with capabilities for handling rich-context scenarios,
we design a novel pipeline that automatically generates instruction-following data, \aaainew{termed \ourinstructdata}, specifically focusing on rich-context descriptions, paired with corresponding grounding masks, also along with hard negative samples.
Second, we enhance the learning of two essential tokens [SEG]/[REJ] to 
resolve inconsistencies between conventional autoregressive training and our primary objective. 
Powered by these innovations, \modelname{} excels in hallucination detection while also significantly outperforming previous state-of-the-art methods in the pixel-level grounding task with over 10\% improvement.

To evaluate \modelname{}’s performance in real-world hallucination detection scenarios, we establish a new benchmark, \ourbenchmark{} (Real response, Rich-context Hallucination Benchmark), for Hallucination Detection. 
As evidenced in the 
Table~\ref{tab:table_1_final}~(Right), 
compared with prior benchmarks~\cite{pope, hu2023ciem, amber, chen2024unified}, ours demonstrates the significant superiority across multiple dimensions. \ourbenchmark{} incorporates real responses from 18 powerful MLLMs spanning diverse model architectures and scales, with high-quality human annotations. 
It also features rich-context descriptions with \samson{significantly longer} response lengths, \samson{as well as} including object-, attribute-, and relationship-level details, thus making it suitable for evaluating both hallucination checkers and the hallucination 
\samson{behaviors}
in MLLMs. 

In summary, our key contributions are as follows:
\aaainew{
\begin{itemize}
    \item We introduce \modelname{}, the first visual-back grounding framework for hallucination detection in MLLMs, which functions efficiently without extra references and experts, processes rich-context responses, and also offers interpretability on both modalities.
    \item We design a novel pipeline for generating instruction-tuning data, \aaainew{named \ourinstructdata}, focusing on rich-context object descriptions, paired with grounding masks, and curated challenging negative samples.
    \item We propose \ourbenchmark{}, a new real-response and rich-context benchmark for hallucination detection, featuring high-quality and rich annotations, and reflecting real-world distributions of hallucination detection tasks.
    \item Our \modelname{}, while providing interpretable ability, not only surpasses existing complex frameworks with state-of-the-art performance on \ourbenchmark{} and POPE, rivaling GPT-4o's hallucination detection capability, but also delivers a 10\% improvement over previous methods in the challenging pixel-level grounding task.
\end{itemize}
}

\vspace{-1mm}
\section{Related Work}

\subsection{Grounding LLM}
While large language models exhibit strong reasoning capabilities in \samson{NLP tasks},
researchers are progressively extending these abilities to the domain of multimodal large language models. Through aligning visual and textual inputs, models~\cite{llava, llava15,zhu2023minigpt} are able to understand visual information, significantly broadening their range of applications. Furthermore, grounding capabilities \samson{have also been} explored in the multimodal large language models, enabling them not only to comprehend an entire image, but also to understand localized information, or even output precise spatial information~\cite{peng2023kosmos,chen2023shikra, zhang2023gpt4roi,lai2024lisa}. Among these, LISA~\cite{lai2024lisa} was the first work to touch the reasoning segmentation task, which \samson{was} later extended to multi-object grounding~\cite{yang2023improved, rasheed2024glamm}, multi-round conversation~\cite{wang2024segllm} and further introduce rejection capabilities~\cite{Xia_2024_CVPR}. However, 
\samson{despite these advances, their ability}
to accurately comprehend and reject rich-context queries remains limited.

\vspace{-0.5mm}

\subsection{Hallucination Detection}
Research on hallucination detection in MLLMs~\cite{bai2024hallucination, liu2024survey} has gained significant attention due to the critical need for reliable multimodal systems. Current approaches can be categorized into reference-based and reference-free methods.
The former compares generated content against ground-truth to identify discrepancies~\cite{wang2023evaluation}, or utilizes external expert models
to extract visual information that serves as reference points~\cite{liu2023mitigating, wang2023amber, chen2024unified}. These approaches face significant limitations in real-world applications due to their dependence on annotated ground truth data and external expert models.
For reference-free method, 
FaithScore~\cite{jing2023faithscore} leverages multimodal question-answering capabilities for reference-free hallucination detection. However, the method's binary response paradigm restricts interpretability and renders the approach susceptible to bias. Moreover, the requirement for external experts to break down inputs into basic units adds computational cost and \samson{hampers} performance on rich-context inputs. In contrast, our proposed method is able to efficiently handle rich-context queries while providing interpretability.

\section{Method}

\subsection{Verification by Visual Back Framework}

\paragraph{Problem Formulation. }
Our work addresses the problem of hallucination detection in Multimodal Large Language Models (MLLMs), which determines whether a given response from MLLMs contains visual hallucinations when conditioned on an input image. Specifically, let an MLLM $\Phi$ generate a descriptive response $R = \{r_1, r_2, \dots, r_n\}$ for an image $I$, where each rich-context sentence $r_i$ describes a specific visual part of the image: $ R = \Phi ( I ) $. 
Sentences $r_i$ may include visual hallucinations that conflict with the actual visual input, as the MLLMs rely on learned patterns rather than adhering to the visual information in $I$.
The goal of hallucination detection is to evaluate each $r_i$ and determine whether it contains a hallucination. This is formulated as a binary classification task for each $r_i$, where the label $y_i \in \{0, 1\}$ indicates the presence or absence of hallucinations. Specifically:
\begin{equation}
    y_i =
\begin{cases} 
0 & \text{if } r_i \text{ aligns with the visual input } I, \\
1 & \text{if } r_i \text{ conflicts with the visual input } I.
\end{cases}
\end{equation}

\begin{figure}[t]
    \centering
    \includegraphics[width=0.9\linewidth]{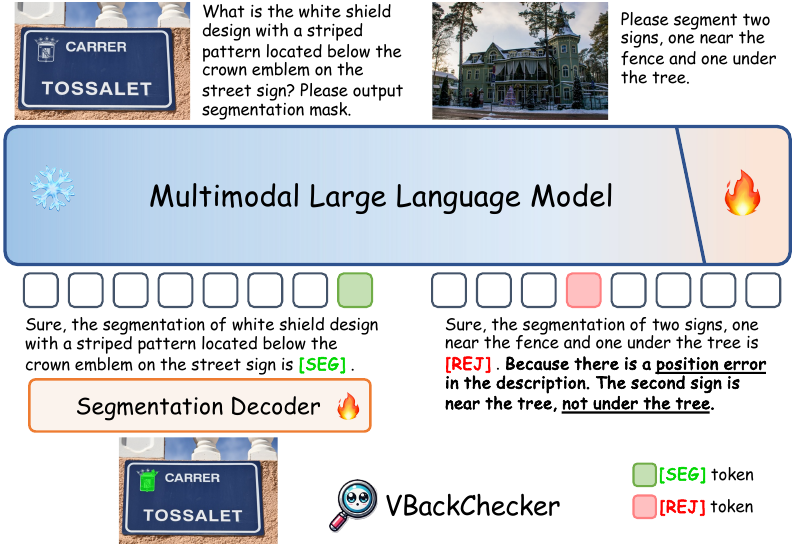}
    \vspace{-1mm}
    \caption{
    The framework of our proposed \modelname{}
    }
    \label{fig:model}
    \vspace{-4mm}
\end{figure}

\begin{figure*}[ht]
    \centering
    \includegraphics[width=0.9\linewidth]{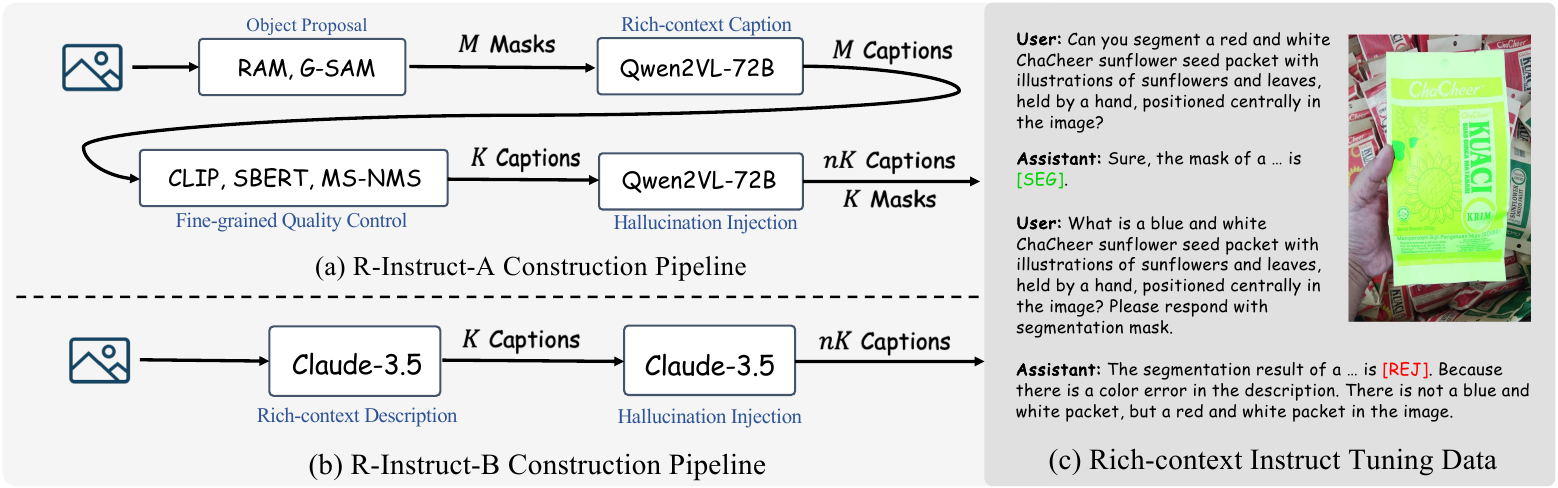}
    \vspace{-1mm}
    \caption{
    The Generation Pipeline of Rich-Context Instruct Tuning Data (\ourinstructdata).
    }
    \label{fig:pipeline}
    \vspace{-4mm}
\end{figure*}

\paragraph{VBackChecker Framework. }
We propose the Verification by Vision Back Framework (\modelname{}), as illustrated in Fig.~\ref{fig:teaser}, to tackle this problem by leveraging a pixel-level Grounding LLM with reasoning and referring segmentation capabilities. The key idea behind \modelname{} is straightforward yet logical: Seeing is Believing. It verifies the response by “looking back” at the visual input.
For each $r_i$ in the MLLM's generated response, \modelname{} determines whether $r_i$ could be accurately grounded back to the input image $I$. If the grounding is successful, the sentence is free of hallucinations ($y_i = 0$); otherwise, it indicates the sentence contains a hallucination ($y_i = 1$).

\modelname{} employs a conversational model to predict either a [SEG] token or a [REJ] token to perform hallucination detection: When there is no hallucination in the MLLM's response $r_i$ based on the input image $I$, \modelname{} outputs a [SEG] token. 
The embedding of [SEG] before the LLM head is passed to a mask decoder (as in LISA~\cite{lai2023lisa}) to decode an object mask $m \in \mathbb{R}^{H \times W}$ corresponding to the described part in $r_i$. When the response $r_i$ contains a hallucination, \modelname{} outputs a [REJ] token and generates a detailed explanation in natural language indicating where $r_i$ conflicts with the visual content of $I$.
By combining pixel-level grounding and language-based reasoning, \modelname{} not only detects hallucinations but also provides interpretable outputs in both visual and language modalities. This dual capability ensures robust and reliable hallucination checking for MLLMs.  

However, existing Grounding LLMs lack \samson{sufficient} discriminative power for rich-context queries. While GSVA~\cite{Xia_2024_CVPR} attempts to reject references to non-existent objects, it struggles with fine-grained understanding. For example, when \samson{querying} about a “... person in white” in an image with a person in black \samson{near} a white wall, GSVA incorrectly predicts [SEG] instead of [REJ], failing to distinguish subtle attributes like object-specific colors. 

\subsection{R-Instruct Data Generation}
\label{sec:instruct_data_generation}

To address this gap, we introduce Rich-context Instruct Tuning data (\ourinstructdata), built by the automated pipeline designed to generate rich-context visual instruction data for grounding and hallucination \samson{detection}. 
R-Instruction includes grounding masks for positive queries and hallucination types with explanations for negative queries, \samson{enabling interpretability in both visual and language modalities}

As illustrated in Figure~\ref{fig:pipeline}(a), we propose an automated four-step procedure to construct our R-Instruct data. 
Let $\mathcal{I}$ be the set of images sampled from the large-scale and high-quality SA1B dataset. For each image $I \in \mathcal{I}$:

\noindent\textbf{Object Proposal.} 
We first employ the Recognize Anything Model~\cite{ram} and the Grounded Segment Anything Model~\cite{gsam} to identify candidate objects $\{\Omega_m\}_{m=1}^{M}$, where each $\Omega_m$ includes a bounding box and the corresponding segmentation mask.

\noindent\textbf{Rich-context Caption.}
For each proposal $\Omega_m$, we employ Qwen2-VL-72B~\cite{wang2024qwen2} to generate a detailed caption $C_m$, capturing the category, shape, attributes (e.g., color, texture), spatial relationships, and distinctive features. This yields a set of object-description pairs, ensuring unique identification of objects within the image.

\noindent\textbf{Quality Control.}
To ensure accurate and diverse captions, by using CLIP~\cite{CLIP} and SBERT~\cite{reimers2019sentence}, we apply a multi-criteria filtering process:
(1) Foreground Background Check. 
Given a proposal {\small $\Omega_m$}, we compute CLIP scores for its foreground-masked image 
{\small $I_{\mathrm{fg}}^m$} 
and background-masked image 
{\small $I_{\mathrm{bg}}^m$}:
{\small $S_{\mathrm{fg}}^m = \mathrm{CLIP}(I_{\mathrm{fg}}^m, C_m)$,  $S_{\mathrm{bg}}^m = \mathrm{CLIP}(I_{\mathrm{bg}}^m, C_m)$}.
We retain captions where 
{\small $S_{\mathrm{fg}}^m > S_{\mathrm{bg}}^m$}
, ensuring the description aligns with the object rather than background artifacts.
(2) Vision-Text Consistency.
Captions with {\small $S_{\mathrm{fg}}^m$} below a confidence threshold (0.5) are discarded, filtering out poorly aligned descriptions.
(3) Multimodal Semantic NMS (Non-Maximum Suppression). 
We propose a Multimodal Semantic NMS to remove near-duplicate proposals. We treat textual similarity $\mathrm{sim}(C_m, C_n)$ (computed via SBERT) as “semantic IoU” and use 
{\small $S_{\mathrm{fg}}^m$}
as a confidence score. Captions with high similarity to a higher-scoring reference are iteratively removed, ensuring each object retains its most coherent description.
This process yields $K$ high-quality samples $(\Omega_k, C_k)$, each with a precise mask and a verified rich-context caption.

\noindent\textbf{Hallucination Injection.}
To create challenging negatives for hallucination detection, we prompt Qwen2-VL to perturb each valid caption $C_k$ using the original image and contextual object descriptions. The model introduces misleading details—e.g., incorrect attributes or nonexistent objects—yielding hallucinated variants. Each is annotated with its hallucination type (e.g., incorrect color, nonexistent object) and an explanation, enabling the model to distinguish accurate from misleading content.

As shown in Figure~\ref{fig:pipeline}(b), we also collect holistic, multi-object captions (without per-object grounding masks) to better mimic real MLLM outputs. Using Claude-3.5, we generate comprehensive descriptions covering all objects in an image as additional positive samples, then apply the same perturbation process to produce negatives. The resulting dataset(\ourinstructdata) includes 30k images, 300k positive samples (100k with masks), and 500k negatives. Data with masks (Figure~\ref{fig:pipeline}(a)) forms R-Instruct-A; data without masks (Figure~\ref{fig:pipeline}(b)) forms R-Instruct-B. Positive responses contain [SEG], while negative ones include [REJ] with hallucination explanations (Figure~\ref{fig:pipeline}(c)).

\subsection{Instruction Tuning}

The training methodology for Rich-context Instruction Tuning leverages a self-regressive learning framework tailored for multimodal grounding and hallucination detection. The overall loss function consists of two components: language loss ( $\mathcal{L}_L$ ) and grounding loss ( $\mathcal{L}_G$ ), defined as:

\begin{equation}
    \mathcal{L} = \mathcal{L}_L(t, t') + \mathcal{L}_G(m, m'),
\end{equation}

\noindent where  $t$  and  $m$  denote the predicted response and predicted mask, respectively, and  $t'$  and  $m'$  represent their corresponding ground truths. 
Specifically, $t = L_{\text{dec}}(h)$ , the textual output decoded by the language decoder  $L_{\text{dec}}$  from the hidden state  $h$.
{\small $m = G_{\text{dec}}(\text{MLP}(h[\text{SEG}]), I)$}, the predicted grounding mask generated by decoding the vision-guided hidden state  {\small $h[\text{SEG}]$}, processed through an MLP projector and conditioned on the image  $I $. $G_{\text{dec}}$ is the Mask Decoder, whose architecture follows that of SAM~\cite{SAM}.

\begin{figure*}[ht]
    \centering
    \includegraphics[width=0.99\linewidth]{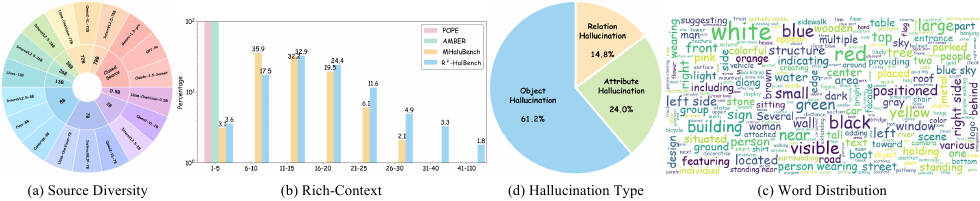}
    \vspace{-2.5mm}
    \caption{
    A comprehensive analysis of our \ourbenchmark{} across four distinct dimensions.
    }
    \label{fig:bench}
    \vspace{-3mm}
\end{figure*}

\begin{figure*}[ht]
    \centering
    \includegraphics[width=0.99\linewidth]{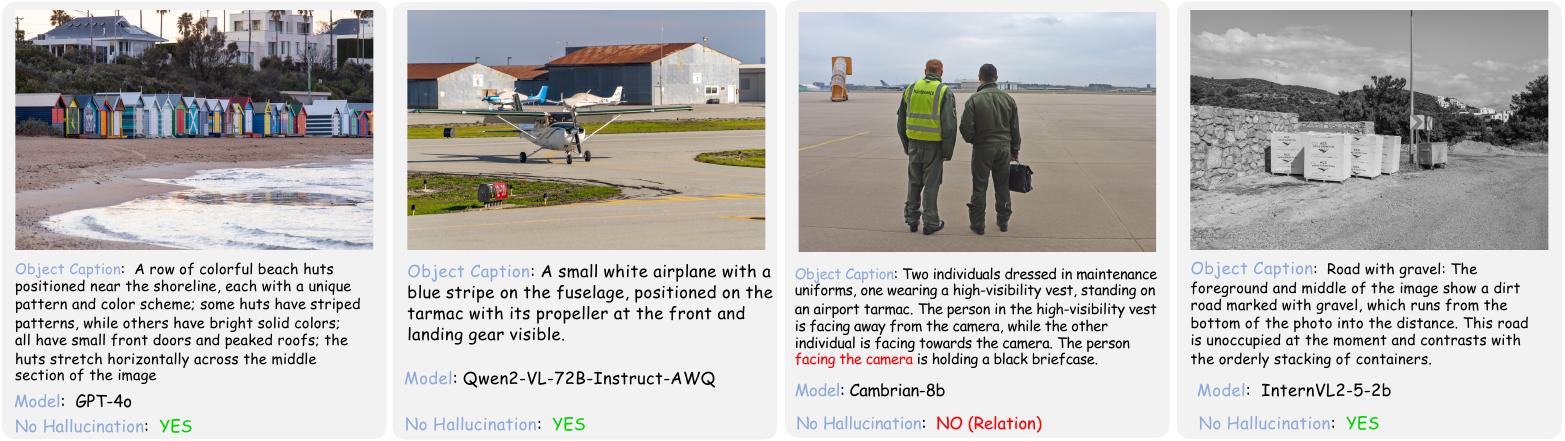}
    \vspace{-2mm}
    \caption{
    Samples of our \ourbenchmark.
    Each rich-context response, generated from various advanced MLLMs, is systematically annotated with corresponding hallucination categories.
    }
    \label{fig:bench_sample}
    \vspace{-5mm}
\end{figure*}

The language loss $\mathcal{L}_L$ focuses solely on the response portion of the output sequence, evaluating the next-token prediction with Cross-Entropy Loss.
The grounding loss $\mathcal{L}_G$ employs a combination of per-pixel Binary Cross-Entropy Loss and DICE Loss to optimize the segmentation mask prediction.
Furthermore, unlike standard MLLM instruction tuning, where every output token equally contributes to the supervision signal, the performance of our grounding LLM, \modelname{}, hinges on its ability to make accurate predictions for the special tokens [SEG] and [REJ]. These tokens represent the model’s decision to either ground or reject a query and are crucial for hallucination detection. To emphasize their importance during training, we enhance their contributions to the loss function by applying higher weights to their Cross-Entropy terms. The modified language loss for these tokens is defined as:

\begin{equation}
    \scriptstyle
    \mathcal{L}_L^{'} = -\sum \alpha_i \log P(t_i | t_{<i}),\; \\
    \alpha_i = 
    \begin{cases}
        \lambda,\! & \text{if } t_i \in \scriptstyle \{\text{[SEG]}, \text{[REJ]} \} \\
        1,\! & \text{otherwise}
    \end{cases}
\end{equation}

\noindent where $i$ indexes tokens in the response. $\lambda>1$ is a hyperparameter that amplifies the contributions of [SEG] and [REJ], guiding the model to emphasize learning critical decisions. 
\section{R²-HalBench}

To \samson{support} reliable evaluation of hallucination detection in rich-context MLLM responses, we introduce R²-HalBench, a benchmark with 3,000 rich-context object descriptions, generated by 18 advanced MLLMs\samson{, from 10,000 images}. Designed to reflect real-world hallucination distributions, it ensures a comprehensive and representative assessment. Figure~\ref{fig:bench_sample} showcases several samples of the benchmark.

\subsection{Dataset Collection and Annotation}
 
R²-HalBench consists entirely of real MLLM outputs, preserving natural hallucination patterns rather than artificially injecting errors by LLMs.
We prompt MLLMs to generate rich-context descriptions for objects in high-quality images from the SA1B dataset~\cite{SAM}, ensuring alignment with real-world hallucination distributions.
Human annotators then verify each description against the corresponding image, classifying samples as “without hallucination” or “with hallucination”. Hallucinated cases are further categorized into three types: Object-level (category/existence), Attribute-level (e.g., color, shape, material), and Relation-level (spatial positioning, interactions). Each sample is reviewed by three independent annotators with at least college-level education, and final labels are determined by majority voting. This rigorous process ensures a high-quality benchmark that accurately reflects real-world hallucination phenomena.

\begin{table*}[htbp]

\centering
\normalsize

\scalebox{0.76}{
\begin{tabular}{cc|ccc|ccc|ccc|ccc}
\toprule

\multirow{3}{*}{\textbf{Method}} &
\multirow{3}{*}{\textbf{LLM}} & 
\multicolumn{9}{c|}{\textbf{gRefCOCO}} &   
\multicolumn{3}{c}{\textbf{R-Instruct-A Val}} \\

\cmidrule{3-9}\cmidrule{10-14}

& & 
\multicolumn{3}{c|}{Validation Set} & 
\multicolumn{3}{c|}{Test Set A} & 
\multicolumn{3}{c|}{Test Set B} & 
\multicolumn{3}{c}{Validation Set} \\

& 
& IoU
& N-acc
& T-acc
& IoU
& N-acc
& T-acc 
& IoU
& N-acc
& T-acc
& IoU
& N-acc
& T-acc \\

\midrule

ReLA~\cite{liu2023gres} & -
& 63.6 & 56.4 & 96.3
& 70.0 & 59.0 & 97.7
& 61.0 & 59.9 & 95.4
& - & - & - \\

LISA-7B~\cite{lai2023lisa} & Vicuna-7B
& 61.6 & 54.7 & -
& 66.3 & 50.0 & -
& 58.4 & 51.9 & -
& - & - & - \\

LISA-13B~\cite{lai2023lisa} & Llama2-13B
& 63.5 & 55.3 & -
& 68.2 & 52.2 & -
& 61.8 & 56.2 & -
& - & - & - \\

SAM4MLLM~\cite{chen2024sam4mllm} & Llama3-8B
& 71.9 & 66.1 &	-
& 74.2 & 63.9 & -
& 65.3 & 60.0 & -
& - & - & - \\

GSVA-7B~\cite{Xia_2024_CVPR} & Vicuna-7B
& 66.5 & 62.4 & -
& 71.1 & 65.3 & -
& 62.2 & 60.6 & -
& 13.2 & 0.4 & 99.8 \\

GSVA-13B~\cite{Xia_2024_CVPR} & Llama2-13B
& 70.0 & 66.0 & -
& 73.3 & 64.7 & -
& 65.5 & 62.5 & -
& 12.7 & 0.3 & 99.8 \\

\midrule

\rowcolor{gray!17} Ours & Vicuna-7B
& 80.2 & 81.3 & 95.2
& 74.0 & 74.3 & 94.6
& 70.8 & 72.8 & 97.3
& 63.8 & 75.3 & 70.7 \\

\bottomrule
\end{tabular}
} 
\vspace{-2mm}
\caption{
Performance of Pixel-level Grouding task on gRefCOCO and R-Instruct-A Val. Metrics include IOU, negative accuracy (N-Acc) and true positive accuracy (T-Acc).
Our \modelname{} surpasses prior methods in pixel-level grounding task.
}
\vspace{-2mm}
\label{tab:result_on_grefcoco_faithseg}
\end{table*}

\begin{figure*}[ht]
    \centering
    \includegraphics[width=0.99\linewidth]{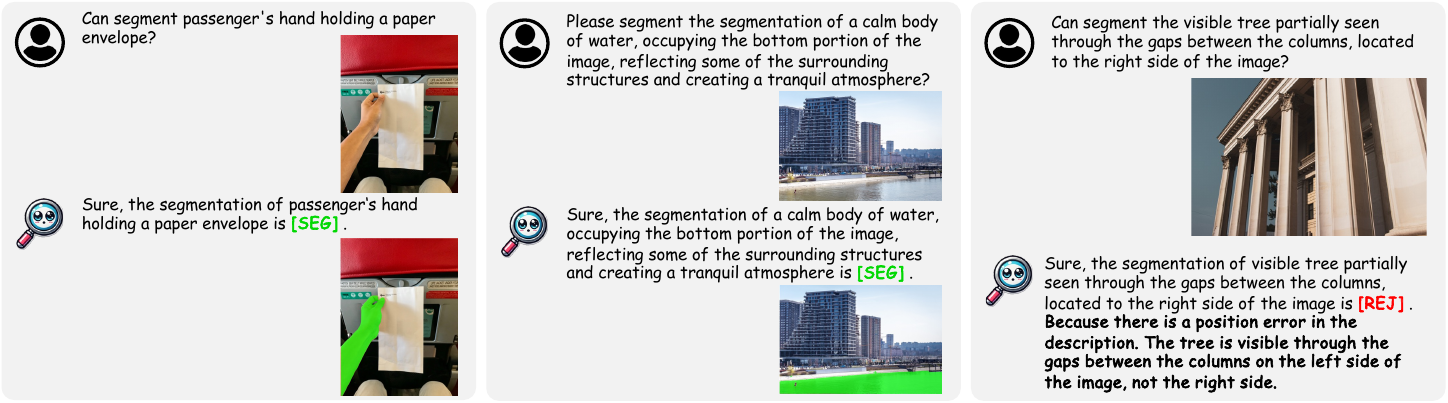}
    \vspace{-2mm}
    \caption{
    Visualization results of our \modelname{}
    which employs a conversational model to predict either a [SEG] token or a [REJ] token to perform hallucination detection. Having a [SEG] token indicates that there is no hallucination, otherwise 
    \modelname{} outputs a [REJ] token with detailed explanation.
    } 
    \label{fig:result_visualization}
    \vspace{-3mm}
\end{figure*}

\begin{table*}[htbp]

\newcommand{\greentick}{\Large\textcolor{green!70!black}{$\checkmark$}}
\newcommand{\redcross}{{\textcolor{red}{$\times$}}}

\centering
\normalsize

\scalebox{0.70}{
\begin{tabular}{cc|cc|cccccc|ccc}
\toprule

\multirow{2}{*}{\textbf{Evaluator}} &
\multirow{2}{*}{\textbf{LLM}} & 
\multicolumn{2}{c|}{\textbf{Explainability}} &   
\multicolumn{6}{c|}{\textbf{R²-HalBench}}  & 
\multicolumn{3}{c}{\textbf{POPE Acc}} \\

\cmidrule{3-4}\cmidrule{5-10}\cmidrule{11-13}

& 
& \textbf{Visual}
& \textbf{Text} 

& \textbf{Acc}
& \textbf{Neg.Acc} 
& Object
& Attribute
& Relation
& \textbf{Pos.Acc} 

& Random
& Popular
& Adversarial \\

\midrule

GPT-4o~\cite{achiam2023gpt} & -  
&  &  
& 63.9 & 31.4 & 37.5 & 23.9 & 18.6 & 93.9
& 96.6 & 87.4 & 85.0 \\

GPT-4o~\cite{achiam2023gpt} & -
&  & \greentick
& 68.3 & 53.8 & 61.6 & 43.2 & 39.2 & 81.6
& 98.8 & 91.4 & 79.9 \\

LLaVA-OneVision~\cite{li2024llava} & 7B
&  &  
& 47.9 & 98.8 & 98.9 & 98.8 & 98.5 & 0.9
& 93.0 & 85.6 & 36.6 \\

LLaVA-OneVision~\cite{li2024llava} & 7B
&  & \greentick
& 48.0 & 99.1 & 98.5 & 99.9 & 99.9 & 0.9
& 99.2 & 89.1 & 34.9 \\

Qwen2VL~\cite{wang2024qwen2} & 7B
&  &  
& 47.7 & 76.4 & 75.6 & 77.3 & 78.4 & 21.2
& 71.9 & 68.2 & 42.7 \\

Qwen2VL~\cite{wang2024qwen2} & 7B
&  & \greentick
& 48.9 & 79.8 & 80.1 & 79.2 & 79.9 & 20.5
& 77.5 & 72.6 & 41.4 \\

Qwen2VL~\cite{wang2024qwen2} & 72B
&  &  
& 48.5 & 83.2 & 83.1 & 86.4 & 78.4 & 16.5
& 64.6 & 62.3 & 44.7 \\

Qwen2VL~\cite{wang2024qwen2} & 72B
&  & \greentick 
& 47.9 & 81.7 & 82.4 & 79.8 & 81.9 & 16.8
& 68.3 & 65.3 & 42.7 \\

\midrule

GSVA~\cite{Xia_2024_CVPR} & 7B
& \greentick &  
& 49.9 & 0.5 & 0.7 & 0.3 & 0.0 & 99.3
& 23.3 & 48.3 & 72.0 \\

GSVA~\cite{Xia_2024_CVPR} & 13B
& \greentick &  
& 49.8 & 0.2 & 0.3 & 0.0 & 0.0 & 99.5
& 0.5 & 10.8 & 64.9 \\

FaithScore & 7B+13B+GPT-3.5
&  &  
& 59.3 & 75.3 & 76.7 & 73.7 & 71.5 & 44.6
& 82.7 & 83.0 & 78.9 \\

\midrule

\rowcolor{gray!17} Ours & 7B
& \greentick & \greentick
& 62.5 & 64.1 & 64.2 & 63.4 & 64.9 & 60.9
& 99.3 & 93.2 & 70.3 \\

\bottomrule

\end{tabular}
} 
\vspace{-2mm}
\caption{
Performance of Hallucination Detection on \ourbenchmark{} and POPE. 
By leveraging the backward grounding,  \modelname{} achieves superior performance and offers interpretability on both \aaainew{modalities}, despite of only 7B parameters.
}
\vspace{-2mm}
\label{tab:results_on_HalMetaBench}
\end{table*}

\subsection{Dataset Statistics}
\ourbenchmark{} is designed to replicate real-world hallucination detection scenarios. Figure~\ref{fig:bench} presents a detailed analysis from four key perspectives:

\newlength{\myBenchmarkSpace}
\setlength{\myBenchmarkSpace}{0.7mm}

\vspace{\myBenchmarkSpace}

\noindent \textbf{(a) \textit{Model Diversity}}
To ensure a broad representation of responses across different model architectures and scales,
18 different MLLMs are employed to generate responses, 
covering both open- and closed-source models, 
and featuring model sizes ranging from 0.5B to 78B parameters.

\vspace{\myBenchmarkSpace}

\noindent \textbf{(b) \textit{Rich-Context Description Length} }
Different from previous benchmarks such as POPE~\cite{pope} and AMBER~\cite{amber}, which rely on simple object name queries and lack rich-context descriptions, \ourbenchmark{} contains diverse and contextually rich descriptions of varying lengths. Even compared to MHaluBench which provides reference-based evaluation, \ourbenchmark{} features longer and more complex descriptions while covering responses from 18 MLLMs—significantly more than 5 MLLMs used in MHaluBench. This reduces evaluation bias and enhances the robustness of hallucination detection assessments.

\vspace{\myBenchmarkSpace}

\noindent \textbf{(c) \textit{Hallucination Categories} }
\ourbenchmark{} systematically \samson{covers} object-, attribute-, and relation-level hallucinations, enabling a thorough evaluation of detection methods and a detailed analysis of how different detectors handle real-world hallucination types in practical applications.

\vspace{\myBenchmarkSpace}

\noindent \textbf{(d) \textit{Lexical Distribution of Rich-Context Descriptions}}
Word frequency distribution analysis in \ourbenchmark{} \samson{reveals a dominance of} attributes and positional terms, effectively capturing fine-grained contextual details. \samson{This makes the benchmark particularly} well-suited for evaluating models that need to discern subtle inaccuracies in descriptions.

\section{Experiments}

In this section, we show comprehensive experiments conducted to evaluate the performance of our proposed \modelname{} on both grounding and hallucination detection tasks, as well as ablation studies to analyze the effectiveness and robustness of our framework. 
\aaainew{The \modelname{} follows LISA in using LLaVa-vicuna-v1.1 and SAM as the base model. It is trained in two stages: the first stage leverages LISA's mixed training dataset along with gRefCOCO, and the second stage uses our proposed \ourinstructdata ~data.}

\begin{table*}[t]

\begin{center}

\scalebox{1}{

\begin{tabular}{@{\hspace{-3mm}}c@{\hspace{4mm}}c@{\hspace{0mm}}}

\begin{minipage}{0.65\textwidth}

\scalebox{0.72}{
\begin{tabular}{cccc|cccc}
\toprule

\textbf{VLM Source} & 
\textbf{Acc} & 
\textbf{Neg.Acc} &  
\textbf{Pos.Acc} &
\textbf{VLM Source} & 
\textbf{Acc} & 
\textbf{Neg.Acc} &  
\textbf{Pos.Acc} \\

\midrule

Claude-3-5-Sonnet & 63.2 & 38.3 & 71.3 
& InternVL-2.5-2B & 61.6 & 77.3 & 45.8 \\

GPT-4o & 54.1 & 56.3 & 52.5 
& InternVL-2.5-8B & 60.2 & 72.7 & 45.8 \\

Gemini-1.5-Pro & 56.2 & 59.6 & 54.7 
& InternVL-2.5-26B & 63.4 & 64.8 & 62.1 \\

Qwen2-VL-2B & 62.8 & 52.9 & 70.5 
& InternVL-2.5-38B & 65.1 & 76.2 & 55.6 \\

Qwen2-VL-7B & 67.9 & 62.3 & 75.9 
& InternVL-2.5-78B & 61.5 & 76.1 & 46.0 \\

Qwen2-VL-72B & 59.5 & 50.8 & 69.6 
& LLaVA-OneVision-0.5B & 59.7 & 55.9 & 64.3 \\

InstructBLIP-7B & 69.4 & 55.0 & 79.3 
& LLaVA-OneVision-7B & 68.4 & 66.0 & 69.8 \\

Cambrain-8B & 57.3 & 56.8 & 58.2 
& LLaVA-OneVision-72B & 63.5 & 53.3 & 72.6 \\

Fuyu-8B & 73.7 & 73.1 & 75.0 
& LLaVa-1.5-13B & 46.2 & 31.8 & 64.7 \\

\bottomrule
\end{tabular}
}
\label{tab:results_on_sources}
\end{minipage}

&
\begin{minipage}{0.3\textwidth}

\scalebox{0.793}{
\begin{tabular}{cccc}
\toprule

\textbf{
\makecell{Context Len \\ (words)}
} & 
\textbf{Acc} & 
\textbf{Neg.Acc} &  
\textbf{Pos.Acc} \\

\midrule

1-5 & 67.3 & 56.5 & 75.9 \\
6-10 & 61.9 & 54.2 & 68.3 \\
11-15 & 61.8 & 63.3 & 60.4 \\
16-20 & 59.1 & 64.4 & 54.6 \\
21-30 & 60.1 & 68.4 & 50.9 \\
31-50 & 63.7 & 67.2 & 59.7 \\
51-110 & 59.1 & 61.5 & 55.6 \\

\bottomrule
\end{tabular}

}
\label{tab:results_on_length}
\end{minipage}

\end{tabular}
}
\end{center}
\vspace{-4mm}
\caption{
\textbf{(Left)} \modelname{}'s evaluation results of samples from different MLLMs in our \ourbenchmark{}.
\textbf{(Right)} Evaluation results of samples with various different length, measured in word count.
Our approach maintains consistency and reliability. 
}
\label{tab:results_on_source_length}
\vspace{-4mm}
\end{table*}

\begin{table}[tbp]

\newlength{\myspace}
\setlength{\myspace}{1mm}

\centering
\small

\scalebox{0.75}{
\begin{tabular}{>{\raggedleft\arraybackslash}l|ccc|ccc}
\toprule

\multirow{2}{*}{\makecell[c]{\hspace{8mm} \vspace{-2.3mm} \textbf{Model}}} &
\multicolumn{3}{c|}{\textbf{R²-HalBench }} &   
\multicolumn{3}{c}{\textbf{POPE}} \\

\cmidrule{2-4}\cmidrule{5-7}

&
\textbf{Acc} & 
\textbf{NegAcc} &
\textbf{PosAcc} &
\textbf{Random} &
\textbf{Popular} &
\textbf{Adv} \\

\midrule

Ours (\modelname) & 
62.5 & 64.1 & 60.9 & 
99.3 & 93.2 & 70.3 \\

\hspace{\myspace} w/o Grounding Loss &
57.3 & 81.3 & 33.4 &
97.6 & 89.5 & 47.2 \\

\hspace{\myspace} w/o SEG/REJ Weight &
59.3 & 55.8 & 62.7 &
99.4 & 92.9 & 67.5 \\

\hspace{\myspace} w/o R-Insturct&
50.4 & 2.3 & 98.6 &
86.3 & 88.9 & 80.3 \\

\hspace{\myspace} \hspace{\myspace} w/o R-Instruct-A &
60.0 & 59.8 & 60.2 &
99.0 & 92.2 & 74.9 \\

\hspace{\myspace} \hspace{\myspace} w/o R-Instruct-B &
58.5 & 53.8 & 63.2 &
99.6 & 93.8 & 60.1 \\

\bottomrule
\end{tabular}
} 
\vspace{-2mm}
\caption{
Ablation studies on different modules.
}
\label{tab:module_ablation}
\vspace{-5mm}
\end{table}

\subsection{Main Results}
\paragraph{Pixel-Level Grounding}
For pixel-level grounding task, we evaluate \modelname{} on two referring segmentation datasets: gRefCOCO~\cite{liu2023gres} and R-Instruct-A-Val\samson{, the validation set of R-Instruct-A}. Following the baseline GSVA, we adopt three metrics: IoU, negative accuracy (N-Acc) which measures rejection accuracy (correctly rejected queries over groundtruth negative samples), and true positive accuracy (T-Acc) which evaluates segmentation accuracy (correctly segmented queries over groundtruth positive samples).
We evaluate \modelname{} on gRefCOCO, which primarily contains simple object references, and on a validation set sampled from the rich-context instruction dataset (Sec~\ref{sec:instruct_data_generation}) to evaluate its ability to handle more complex queries. \modelname{} shows significant improvements over the previous SOTA method in IoU and rejection accuracy on gRefCOCO (Table~\ref{tab:result_on_grefcoco_faithseg}), achieving 95.2\% segmentation accuracy and 81.3\% rejection accuracy, demonstrating its effectiveness as a hallucination checker. Additionally, evaluation on R-Instruct-A-Val, a benchmark designed for rich-context queries, shows \modelname{}’s superior performance compared to GSVA, which struggles to distinguish hallucinated inputs. Despite the difficulty of benchmark, \modelname{} achieves 75.3\% rejection accuracy, confirming its reliability as a hallucination detector.

\paragraph{Hallucination Detection}
For hallucination detection, \modelname{} is evaluated on the proposed \ourbenchmark{} which closely mirrors real-world distributions. 
Figure~\ref{fig:result_visualization} also shows the detailed visualization results of our framework.
The overall accuracy is utilized as the primary metric, along with negative and positive accuracy. 
In addition, results across different hallucination types, including object-, attribute-, and relation-level hallucinations, are presented in Table~\ref{tab:results_on_HalMetaBench} to enable a more detailed analysis.

We compare \modelname{} with several reference-free baselines, including reference-free hallucination checkers like FaithScore, grounding-based methods like GSVA, and advanced MLLMs (both open-source and closed-source) that predict hallucination presence directly from an image and query. To \samson{address} MLLMs’ hallucination issues, we \samson{design} two experimental settings: one where the model explains its decision and the other that provides a binary hallucination prediction. 

Despite having only 7B parameters, \modelname{} outperforms all open-source baselines, including FaithScore which employs a more complex pipeline with LLaMA-7B, LLaVA-13B, and ChatGPT-3.5. Moreover, \modelname{} achieves state-of-the-art performance among open-source models with a simpler end-to-end framework, approaching the performance of GPT-4o. 
\modelname{} is also evaluated on the POPE benchmark 
\samson{, which consists solely of simple} 
object queries. Even in this relatively less challenging setting for hallucination detection, it consistently demonstrates superior performance, further highlighting its strong generalization capability across different hallucination detection tasks.
These results showcase our \modelname{} as a highly effective hallucination checker, offering an efficient, and accurate solution for multimodal applications.

\subsection{Ablation Studies}
We conduct ablation studies to verify the effectiveness of our proposed method. 
As shown in Table~\ref{tab:module_ablation} (Row 2), we establish a controlled setting by reducing our approach to an instruction-tuned LLaVA model trained on the proposed dataset
. This ablation isolates the grounding effect, yielding a significant 5.2\% improvement on \ourbenchmark{}, underscoring its critical role in hallucination detection.
Row 3 demonstrates the impact of emphasizing [SEG] and [REJ] token learning. 
This targeted weighting mechanism that emphasizes key tokens helps the model better distinguish between grounded and hallucinated responses.
Rows 4-6 provide evidence for the effectiveness of our instruction data, particularly rich-context descriptions with and without segmentation masks. These findings underscore the significant contribution of structured training data toward enhancing hallucination detection capability.

To evaluate \modelname{}'s generalization, we examined its performance across various source models, as documented in Table~\ref{tab:results_on_source_length} (Left). Results indicate that \modelname{} achieves consistent hallucination detection performance across diverse MLLM-generated responses, demonstrating robustness and adaptability.
We also analyze effectiveness on responses of varying lengths (1–110 words). Findings in Table~\ref{tab:results_on_source_length} (Right) confirm that \modelname{} remains reliable across diverse lengths, including extensive rich-context responses (31–110 words), further validating its applicability to complex multimodal reasoning tasks.

\vspace{-1mm}
\section{Conclusion}
\vspace{-0.5mm}

\aaainew{With the principle of “Seeing is Believing,” we designed \modelname{}, a novel reference-free hallucination detection framework that handles rich-context responses and offers interpretability through grounding back, without relying on external references or experts. In support of this, we presented an innovative pipeline to generate \aaainew{instruction-tuning} data (\ourinstructdata)}. We also established \ourbenchmark{}, a new benchmark reflecting real-world hallucination detection challenges. 
Experimental results demonstrated that \modelname{} achieves state-of-the-art performance on \ourbenchmark{}, rivaling GPT-4o's capabilities, while substantially improving pixel-level grounding accuracy. 
We expect these contributions to advance development of more reliable multimodal systems for real-world deployment.

\bibliography{aaai2026}

\begin{thebibliography}{37}
\providecommand{\natexlab}[1]{#1}

\bibitem[{Achiam et~al.(2023)Achiam, Adler, Agarwal, Ahmad, Akkaya, Aleman, Almeida, Altenschmidt, Altman, Anadkat et~al.}]{achiam2023gpt}
Achiam, J.; Adler, S.; Agarwal, S.; Ahmad, L.; Akkaya, I.; Aleman, F.~L.; Almeida, D.; Altenschmidt, J.; Altman, S.; Anadkat, S.; et~al. 2023.
\newblock Gpt-4 technical report.
\newblock \emph{arXiv preprint arXiv:2303.08774}.

\bibitem[{Alayrac et~al.(2022)Alayrac, Donahue, Luc, Miech, Barr, Hasson, Lenc, Mensch, Millican, Reynolds et~al.}]{alayrac2022flamingo}
Alayrac, J.-B.; Donahue, J.; Luc, P.; Miech, A.; Barr, I.; Hasson, Y.; Lenc, K.; Mensch, A.; Millican, K.; Reynolds, M.; et~al. 2022.
\newblock Flamingo: a visual language model for few-shot learning.
\newblock \emph{Advances in neural information processing systems}, 35: 23716--23736.

\bibitem[{Bai et~al.(2024)Bai, Wang, Xiao, He, Han, Zhang, and Shou}]{bai2024hallucination}
Bai, Z.; Wang, P.; Xiao, T.; He, T.; Han, Z.; Zhang, Z.; and Shou, M.~Z. 2024.
\newblock Hallucination of multimodal large language models: A survey.
\newblock \emph{arXiv preprint arXiv:2404.18930}.

\bibitem[{Chen et~al.(2023)Chen, Zhang, Zeng, Zhang, Zhu, and Zhao}]{chen2023shikra}
Chen, K.; Zhang, Z.; Zeng, W.; Zhang, R.; Zhu, F.; and Zhao, R. 2023.
\newblock Shikra: Unleashing Multimodal LLM's Referential Dialogue Magic.
\newblock \emph{arXiv preprint arXiv:2306.15195}.

\bibitem[{Chen et~al.(2024{\natexlab{a}})Chen, Wang, Xue, Zhang, Yang, Li, Shen, Liang, Gu, and Chen}]{chen2024unified}
Chen, X.; Wang, C.; Xue, Y.; Zhang, N.; Yang, X.; Li, Q.; Shen, Y.; Liang, L.; Gu, J.; and Chen, H. 2024{\natexlab{a}}.
\newblock Unified hallucination detection for multimodal large language models.
\newblock \emph{arXiv preprint arXiv:2402.03190}.

\bibitem[{Chen et~al.(2024{\natexlab{b}})Chen, Li, Sun, Wang, and Chen}]{chen2024sam4mllm}
Chen, Y.-C.; Li, W.-H.; Sun, C.; Wang, Y.-C.~F.; and Chen, C.-S. 2024{\natexlab{b}}.
\newblock SAM4MLLM: Enhance Multi-Modal Large Language Model for Referring Expression Segmentation.
\newblock In \emph{European Conference on Computer Vision}, 323--340. Springer.

\bibitem[{Chen et~al.(2024{\natexlab{c}})Chen, Wu, Wang, Su, Chen, Xing, Zhong, Zhang, Zhu, Lu et~al.}]{chen2024internvl}
Chen, Z.; Wu, J.; Wang, W.; Su, W.; Chen, G.; Xing, S.; Zhong, M.; Zhang, Q.; Zhu, X.; Lu, L.; et~al. 2024{\natexlab{c}}.
\newblock Internvl: Scaling up vision foundation models and aligning for generic visual-linguistic tasks.
\newblock In \emph{Proceedings of the IEEE/CVF conference on computer vision and pattern recognition}, 24185--24198.

\bibitem[{Hu et~al.(2023)Hu, Zhang, Zhao, and Sun}]{hu2023ciem}
Hu, H.; Zhang, J.; Zhao, M.; and Sun, Z. 2023.
\newblock Ciem: Contrastive instruction evaluation method for better instruction tuning.
\newblock \emph{arXiv preprint arXiv:2309.02301}.

\bibitem[{Jing et~al.(2023)Jing, Li, Chen, Jia, and Du}]{jing2023faithscore}
Jing, L.; Li, R.; Chen, Y.; Jia, M.; and Du, X. 2023.
\newblock FAITHSCORE: Evaluating Hallucinations in Large Vision-Language Models.
\newblock \emph{arXiv preprint arXiv:2311.01477}.

\bibitem[{Kirillov et~al.(2023)Kirillov, Mintun, Ravi, Mao, Rolland, Gustafson, Xiao, Whitehead, Berg, Lo et~al.}]{SAM}
Kirillov, A.; Mintun, E.; Ravi, N.; Mao, H.; Rolland, C.; Gustafson, L.; Xiao, T.; Whitehead, S.; Berg, A.~C.; Lo, W.-Y.; et~al. 2023.
\newblock Segment anything.
\newblock In \emph{Proceedings of the IEEE/CVF International Conference on Computer Vision}, 4015--4026.

\bibitem[{Lai et~al.(2023)Lai, Tian, Chen, Li, Yuan, Liu, and Jia}]{lai2023lisa}
Lai, X.; Tian, Z.; Chen, Y.; Li, Y.; Yuan, Y.; Liu, S.; and Jia, J. 2023.
\newblock Lisa: Reasoning segmentation via large language model.
\newblock \emph{arXiv preprint arXiv:2308.00692}.

\bibitem[{Lai et~al.(2024)Lai, Tian, Chen, Li, Yuan, Liu, and Jia}]{lai2024lisa}
Lai, X.; Tian, Z.; Chen, Y.; Li, Y.; Yuan, Y.; Liu, S.; and Jia, J. 2024.
\newblock Lisa: Reasoning segmentation via large language model.
\newblock In \emph{Proceedings of the IEEE/CVF Conference on Computer Vision and Pattern Recognition}, 9579--9589.

\bibitem[{Li et~al.(2024)Li, Zhang, Guo, Zhang, Li, Zhang, Zhang, Zhang, Li, Liu et~al.}]{li2024llava}
Li, B.; Zhang, Y.; Guo, D.; Zhang, R.; Li, F.; Zhang, H.; Zhang, K.; Zhang, P.; Li, Y.; Liu, Z.; et~al. 2024.
\newblock Llava-onevision: Easy visual task transfer.
\newblock \emph{arXiv preprint arXiv:2408.03326}.

\bibitem[{Li et~al.(2023)Li, Du, Zhou, Wang, Zhao, and Wen}]{pope}
Li, Y.; Du, Y.; Zhou, K.; Wang, J.; Zhao, W.~X.; and Wen, J.-R. 2023.
\newblock Evaluating object hallucination in large vision-language models.
\newblock \emph{arXiv preprint arXiv:2305.10355}.

\bibitem[{Liu, Ding, and Jiang(2023)}]{liu2023gres}
Liu, C.; Ding, H.; and Jiang, X. 2023.
\newblock Gres: Generalized referring expression segmentation.
\newblock In \emph{Proceedings of the IEEE/CVF conference on computer vision and pattern recognition}, 23592--23601.

\bibitem[{Liu et~al.(2023{\natexlab{a}})Liu, Lin, Li, Wang, Yacoob, and Wang}]{liu2023mitigating}
Liu, F.; Lin, K.; Li, L.; Wang, J.; Yacoob, Y.; and Wang, L. 2023{\natexlab{a}}.
\newblock Mitigating hallucination in large multi-modal models via robust instruction tuning.
\newblock \emph{arXiv preprint arXiv:2306.14565}.

\bibitem[{Liu et~al.(2023{\natexlab{b}})Liu, Li, Li, and Lee}]{llava15}
Liu, H.; Li, C.; Li, Y.; and Lee, Y.~J. 2023{\natexlab{b}}.
\newblock Improved baselines with visual instruction tuning.
\newblock \emph{arXiv preprint arXiv:2310.03744}.

\bibitem[{Liu et~al.(2023{\natexlab{c}})Liu, Li, Wu, and Lee}]{liu2023visual}
Liu, H.; Li, C.; Wu, Q.; and Lee, Y.~J. 2023{\natexlab{c}}.
\newblock Visual instruction tuning.
\newblock \emph{Advances in neural information processing systems}, 36: 34892--34916.

\bibitem[{Liu et~al.(2023{\natexlab{d}})Liu, Li, Wu, and Lee}]{llava}
Liu, H.; Li, C.; Wu, Q.; and Lee, Y.~J. 2023{\natexlab{d}}.
\newblock Visual instruction tuning.
\newblock \emph{arXiv preprint arXiv:2304.08485}.

\bibitem[{Liu et~al.(2024)Liu, Xue, Chen, Chen, Zhao, Wang, Hou, Li, and Peng}]{liu2024survey}
Liu, H.; Xue, W.; Chen, Y.; Chen, D.; Zhao, X.; Wang, K.; Hou, L.; Li, R.; and Peng, W. 2024.
\newblock A survey on hallucination in large vision-language models.
\newblock \emph{arXiv preprint arXiv:2402.00253}.

\bibitem[{Panagopoulou et~al.(2024)Panagopoulou, Xue, Yu, Li, Li, Joty, Xu, Savarese, Xiong, and Niebles}]{panagopoulou2024x}
Panagopoulou, A.; Xue, L.; Yu, N.; Li, J.; Li, D.; Joty, S.; Xu, R.; Savarese, S.; Xiong, C.; and Niebles, J.~C. 2024.
\newblock X-InstructBLIP: A Framework for Aligning Image, 3D, Audio, Video to LLMs and its Emergent Cross-Modal Reasoning.
\newblock In \emph{European Conference on Computer Vision}, 177--197. Springer.

\bibitem[{Peng et~al.(2023)Peng, Wang, Dong, Hao, Huang, Ma, and Wei}]{peng2023kosmos}
Peng, Z.; Wang, W.; Dong, L.; Hao, Y.; Huang, S.; Ma, S.; and Wei, F. 2023.
\newblock Kosmos-2: Grounding multimodal large language models to the world.
\newblock \emph{arXiv preprint arXiv:2306.14824}.

\bibitem[{Radford et~al.(2021)Radford, Kim, Hallacy, Ramesh, Goh, Agarwal, Sastry, Askell, Mishkin, Clark et~al.}]{CLIP}
Radford, A.; Kim, J.~W.; Hallacy, C.; Ramesh, A.; Goh, G.; Agarwal, S.; Sastry, G.; Askell, A.; Mishkin, P.; Clark, J.; et~al. 2021.
\newblock Learning transferable visual models from natural language supervision.
\newblock In \emph{International conference on machine learning}, 8748--8763. PMLR.

\bibitem[{Rasheed et~al.(2024)Rasheed, Maaz, Shaji, Shaker, Khan, Cholakkal, Anwer, Xing, Yang, and Khan}]{rasheed2024glamm}
Rasheed, H.; Maaz, M.; Shaji, S.; Shaker, A.; Khan, S.; Cholakkal, H.; Anwer, R.~M.; Xing, E.; Yang, M.-H.; and Khan, F.~S. 2024.
\newblock Glamm: Pixel grounding large multimodal model.
\newblock In \emph{Proceedings of the IEEE/CVF Conference on Computer Vision and Pattern Recognition}, 13009--13018.

\bibitem[{Reimers and Gurevych(2019)}]{reimers2019sentence}
Reimers, N.; and Gurevych, I. 2019.
\newblock Sentence-bert: Sentence embeddings using siamese bert-networks.
\newblock \emph{arXiv preprint arXiv:1908.10084}.

\bibitem[{Ren et~al.(2024)Ren, Liu, Zeng, Lin, Li, Cao, Chen, Huang, Chen, Yan, Zeng, Zhang, Li, Yang, Li, Jiang, and Zhang}]{gsam}
Ren, T.; Liu, S.; Zeng, A.; Lin, J.; Li, K.; Cao, H.; Chen, J.; Huang, X.; Chen, Y.; Yan, F.; Zeng, Z.; Zhang, H.; Li, F.; Yang, J.; Li, H.; Jiang, Q.; and Zhang, L. 2024.
\newblock Grounded SAM: Assembling Open-World Models for Diverse Visual Tasks.
\newblock arXiv:2401.14159.

\bibitem[{Wang et~al.(2023{\natexlab{a}})Wang, Wang, Xu, Zhang, Gu, Jia, Wang, Xu, Yan, Zhang et~al.}]{wang2023amber}
Wang, J.; Wang, Y.; Xu, G.; Zhang, J.; Gu, Y.; Jia, H.; Wang, J.; Xu, H.; Yan, M.; Zhang, J.; et~al. 2023{\natexlab{a}}.
\newblock Amber: An llm-free multi-dimensional benchmark for mllms hallucination evaluation.
\newblock \emph{arXiv preprint arXiv:2311.07397}.

\bibitem[{Wang et~al.(2023{\natexlab{b}})Wang, Wang, Xu, Zhang, Gu, Jia, Yan, Zhang, and Sang}]{amber}
Wang, J.; Wang, Y.; Xu, G.; Zhang, J.; Gu, Y.; Jia, H.; Yan, M.; Zhang, J.; and Sang, J. 2023{\natexlab{b}}.
\newblock An llm-free multi-dimensional benchmark for mllms hallucination evaluation.
\newblock \emph{arXiv preprint arXiv:2311.07397}.

\bibitem[{Wang et~al.(2023{\natexlab{c}})Wang, Zhou, Xu, Shi, Zhao, Xu, Ye, Yan, Zhang, Zhu et~al.}]{haelm}
Wang, J.; Zhou, Y.; Xu, G.; Shi, P.; Zhao, C.; Xu, H.; Ye, Q.; Yan, M.; Zhang, J.; Zhu, J.; et~al. 2023{\natexlab{c}}.
\newblock Evaluation and analysis of hallucination in large vision-language models.
\newblock \emph{arXiv preprint arXiv:2308.15126}.

\bibitem[{Wang et~al.(2023{\natexlab{d}})Wang, Zhou, Xu, Shi, Zhao, Xu, Ye, Yan, Zhang, Zhu et~al.}]{wang2023evaluation}
Wang, J.; Zhou, Y.; Xu, G.; Shi, P.; Zhao, C.; Xu, H.; Ye, Q.; Yan, M.; Zhang, J.; Zhu, J.; et~al. 2023{\natexlab{d}}.
\newblock Evaluation and analysis of hallucination in large vision-language models.
\newblock \emph{arXiv preprint arXiv:2308.15126}.

\bibitem[{Wang et~al.(2024{\natexlab{a}})Wang, Bai, Tan, Wang, Fan, Bai, Chen, Liu, Wang, Ge et~al.}]{wang2024qwen2}
Wang, P.; Bai, S.; Tan, S.; Wang, S.; Fan, Z.; Bai, J.; Chen, K.; Liu, X.; Wang, J.; Ge, W.; et~al. 2024{\natexlab{a}}.
\newblock Qwen2-vl: Enhancing vision-language model's perception of the world at any resolution.
\newblock \emph{arXiv preprint arXiv:2409.12191}.

\bibitem[{Wang et~al.(2024{\natexlab{b}})Wang, Zhang, Li, Kallidromitis, Li, Kato, Kozuka, and Darrell}]{wang2024segllm}
Wang, X.; Zhang, S.; Li, S.; Kallidromitis, K.; Li, K.; Kato, Y.; Kozuka, K.; and Darrell, T. 2024{\natexlab{b}}.
\newblock SegLLM: Multi-round Reasoning Segmentation.
\newblock \emph{arXiv preprint arXiv:2410.18923}.

\bibitem[{Xia et~al.(2024)Xia, Han, Han, Pan, Song, and Huang}]{Xia_2024_CVPR}
Xia, Z.; Han, D.; Han, Y.; Pan, X.; Song, S.; and Huang, G. 2024.
\newblock GSVA: Generalized Segmentation via Multimodal Large Language Models.
\newblock In \emph{Proceedings of the IEEE/CVF Conference on Computer Vision and Pattern Recognition (CVPR)}, 3858--3869.

\bibitem[{Yang et~al.(2023)Yang, Qu, Lai, Tian, Peng, Liu, and Jia}]{yang2023improved}
Yang, S.; Qu, T.; Lai, X.; Tian, Z.; Peng, B.; Liu, S.; and Jia, J. 2023.
\newblock An improved baseline for reasoning segmentation with large language model.
\newblock \emph{arXiv e-prints}, arXiv--2312.

\bibitem[{Zhang et~al.(2023)Zhang, Sun, Chen, Xiao, Shao, Zhang, Chen, and Luo}]{zhang2023gpt4roi}
Zhang, S.; Sun, P.; Chen, S.; Xiao, M.; Shao, W.; Zhang, W.; Chen, K.; and Luo, P. 2023.
\newblock Gpt4roi: Instruction tuning large language model on region-of-interest.
\newblock \emph{arXiv preprint arXiv:2307.03601}.

\bibitem[{Zhang et~al.(2024)Zhang, Huang, Ma, Li, Luo, Xie, Qin, Luo, Li, Liu et~al.}]{ram}
Zhang, Y.; Huang, X.; Ma, J.; Li, Z.; Luo, Z.; Xie, Y.; Qin, Y.; Luo, T.; Li, Y.; Liu, S.; et~al. 2024.
\newblock Recognize anything: A strong image tagging model.
\newblock In \emph{Proceedings of the IEEE/CVF Conference on Computer Vision and Pattern Recognition}, 1724--1732.

\bibitem[{Zhu et~al.(2023)Zhu, Chen, Shen, Li, and Elhoseiny}]{zhu2023minigpt}
Zhu, D.; Chen, J.; Shen, X.; Li, X.; and Elhoseiny, M. 2023.
\newblock Minigpt-4: Enhancing vision-language understanding with advanced large language models.
\newblock \emph{arXiv preprint arXiv:2304.10592}.

\end{thebibliography}

\end{document}